\documentclass{article}
\usepackage{acra}
\usepackage{graphicx}
\usepackage{caption}
\usepackage[boxruled,titlenumbered]{algorithm2e}
\usepackage{amsmath}
\usepackage{amssymb}
\usepackage{subcaption}
\usepackage{adjustbox}
\usepackage[font=small,labelfont=bf]{caption}
\usepackage[spaces,hyphens]{url}

\title{A Low Cost Bin Picking Solution for E-Commerce Warehouse Fulfillment Centers}
\author{Avnish Gupta, Akash Jadhav and Pradyot VN Korupolu \\ Addverb Technologies Pvt. Ltd., India \\
\{avnish.gupta, akash.jhadav, pradyot.korupolu\}@addverb.in }

\begin{document}

\maketitle

\begin{abstract} 
In recent years, the throughput requirements of e-commerce fulfillment warehouses have seen a steep increase. This has resulted in various automation solutions being developed for item picking and movement. In this paper, we address the problem of manipulators picking heterogeneous items placed randomly in a bin. Traditional solutions require that the items be picked to be placed in an orderly manner in the bin and that the exact dimensions of the items be known beforehand. Such solutions do not perform well in the real world since the items in a bin are seldom placed in an orderly manner and new products are added almost every day by e-commerce suppliers. We propose a cost-effective solution that handles both the aforementioned challenges. Our solution comprises of a dual sensor system comprising of a regular RGB camera and a 3D ToF depth sensor. We propose a novel algorithm that fuses data from both these sensors to improve object segmentation while maintaining the accuracy of pose estimation, especially in occluded environments and tightly packed bins. We experimentally verify the performance of our system by picking boxes using an ABB IRB 1200 robot. We also show that our system maintains a high level of accuracy in pose estimation that is independent of the dimensions of the box, texture, occlusion or orientation.  We further show that our system is computationally less expensive and maintains a consistent detection time of 1 second. We also discuss how this approach can be easily extended to objects of all shapes.
\end{abstract}

\section{Introduction}

\noindent Industrial robots used for tasks like assembly operations or pick and place operations can work only on objects whose location and orientation are predefined. These robots without vision system hence are used only for repetitive tasks. This design of a blind robot system increases the cost of automation considering the requirement of expensive object feeding conveyor systems.

\noindent An efficient solution to this situation can be provided by considering a manual assembly environment that has several objects placed randomly in bins around the work area from which required objects can be picked for assembly. Automation of this process requires developing an intelligent robot capable of recognizing and handling objects in bins. This problem, called bin picking, thus consists of picking an object from a group of similar (homogeneous) or different (heterogeneous) objects. The objects in the bin are generally in random orientation and occlude each other; this poses a great challenge to the vision sensors in segmenting the objects to be picked.  

\noindent A great deal of previous research on vision-based bin picking has been limited to recognition of objects and STL matching where the object is predefined due to the complexity involved in segmenting the bin images. In recent years, much work has been proposed on the recognition of objects for the bin picking problem. Some techniques are limited to detection of the objects while others concentrate on picking an object and analyzing the object for pose determination and recognition using the object database \cite{Takenouchi-et-al}, \cite{Ohba}, \cite{Yanagihara}.

\begin{figure}
	\begin{subfigure}{.153\textwidth}
	\centering
	\includegraphics[width=1.0\linewidth]{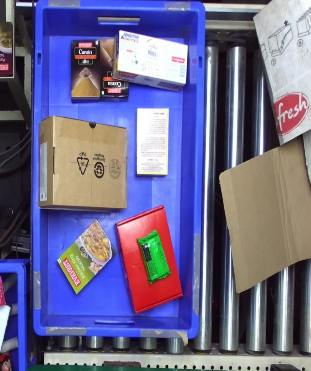}
	\caption{RGB Image}
	\end{subfigure}
	\begin{subfigure}{.153\textwidth}
	\centering
	\includegraphics[width=1.0\linewidth]{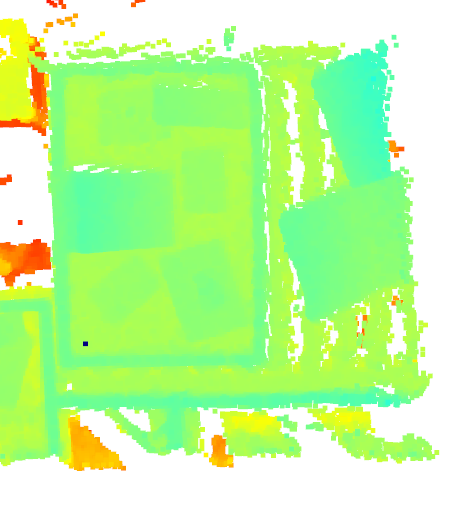}
	\caption{Point Cloud}
	\end{subfigure}
	\begin{subfigure}{.135\textwidth}
	\centering
	\includegraphics[width=1.0\linewidth]{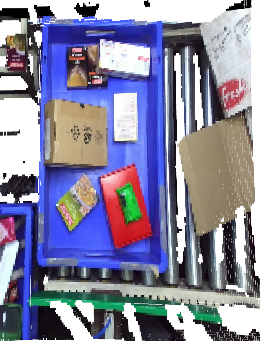}
	\caption{Fused Data}
	\end{subfigure}
	\caption{Illustration of data fusion of RGB Image and Pointcloud}
\end{figure}

\noindent Image segmentation is another well-researched area that involves the division of an image into meaningful sub-regions. If all the objects are segmented individually, then identifying them is relatively easy; however, when the objects are partially occluded more than one object will be considered as a single object and thus making it difficult to identify each one of them. In contrast to the approaches mentioned above, our algorithm attempts to provide a solution for segmentation and localization of boxes of different sizes in a partially occluded bin using a dual vision system based approach. In the dual vision system, the RGB image is fused with a 3D point cloud as shown in Figure 1. Our algorithm proves to work efficiently on textured and non-textured boxes of random sizes in cluttered environments without the requirement of an object database. The accuracy of the algorithm is further evaluated on various test cases which confirms the robustness of the system. The following section compares some of the previous approaches to solving this problem to the proposed algorithm.

\section{Related Work} 

\noindent There has been an immense increase in the demand for a bin picking solution for the past few years. Researchers have proposed various approaches to tackling the bin packing problem. We have discussed some of the approaches that built motivation for our work.

\noindent The most known solution to this problem is STL or CAD matching. Many industrial robots are utilized to pick objects of the same type whose CAD file is already known to the perception system. Using these CAD files, the perception system accurately recognizes the shape and grasp point. \cite{Martin-et-al} demonstrated a flexible assembly of industrial parts where stereo matching and structured light were used for the picking process while CAD models were used to determine the orientation of the objects. Another similar research by \cite{Yang-et-al} uses the Interactive Closest Point algorithm(ICP) on 3D range data for finding the best fitting cloud point with respect to the reference CAD file. These types of instance recognition solutions are limited to organized environments and are dependent on sensor accuracy. They also pose a major issue of maintaining the object CAD database for all the objects which we need to pick.

\noindent Research has also focused on matching extracted 2D features with geometries of known 3D objects using chamfer matching technique \cite{Liu-et-al}. One approach discusses Oriented Chamfer Matching (OCM) on an object database containing 360 images of one object each one representing an orientation at equal intervals created using CAD file of an object \cite{Kuo-et-al}. These solutions have a satisfactory performance in the case of unoccluded and uncluttered objects in the bin and are not quite robust \cite{Kuo-et-al}.

\noindent Another approach to tackle the problem of cluttered object heap is the use of Convolutional Neural Networks (CNNs) trained on a large dataset of images to recognize the object \cite{Bousmalis-et-al}. With the development of 3D RGB-D sensors, there was an increase in the point cloud data. This motivated researchers to further use CNNs for the segmentation of object classes directly from the point cloud of the scene. \cite{Gualtieri-et-al} and \cite{Viereck-et-al} trained CNNs on 3D dense point clouds from synthetic datasets to predict antipodal grasps. These methods have a very good performance if the data is collected from various viewpoints and is large enough so that the algorithm can generalize well. The drawbacks of these methods are the requirements of a powerful GPU system and several months taken in the collection of datasets.

\noindent The major issue with all of the above-mentioned approaches is scalability for a large number of classes of objects. \cite{Mahler and Goldberg} have discussed an approach using Reinforcement Learning to solve the problem of scalability. Our approach addresses the similar problem of scalability but is rather a computationally less intensive and cost-effective solution. 

\noindent The objective of this research is to develop a robust vision system for a manipulator robot to perform automated bin-picking operations generalized for boxes of any sizes and textured surfaces in cluttered environments. The task of bin picking is split into two subtasks, namely box segmentation, and box localization. The two subtasks and the experimental setup is detailed in Section 3, which describes the proposed segmentation algorithm using robust edge detection and elaborates on the box localization process to obtain 6DOF of the box detected. Section 4 discusses the experimental results obtained from the proposed algorithm and real-time bin picking experiments conducted on an ABB IRB 1200 robot. Finally, section 5 and section 6 presents the conclusion and future work. This paper presents a 3D vision based bin-picking for cuboidal objects but this can be further extended for randomly shaped objects. 

\section{Methodology}

\noindent The proposed dual vision system consists of a monocular RGB camera with a 3D depth sensor that provides an organized point cloud. A calibration technique using perspective transform [16] is implemented to fuse the RGB image from the monocular camera with the organized point cloud of the 3D depth sensor such that for every point in the point cloud we have a specific RGB value for it. 

\begin{figure}[t]
\centering
\includegraphics[scale=0.18]{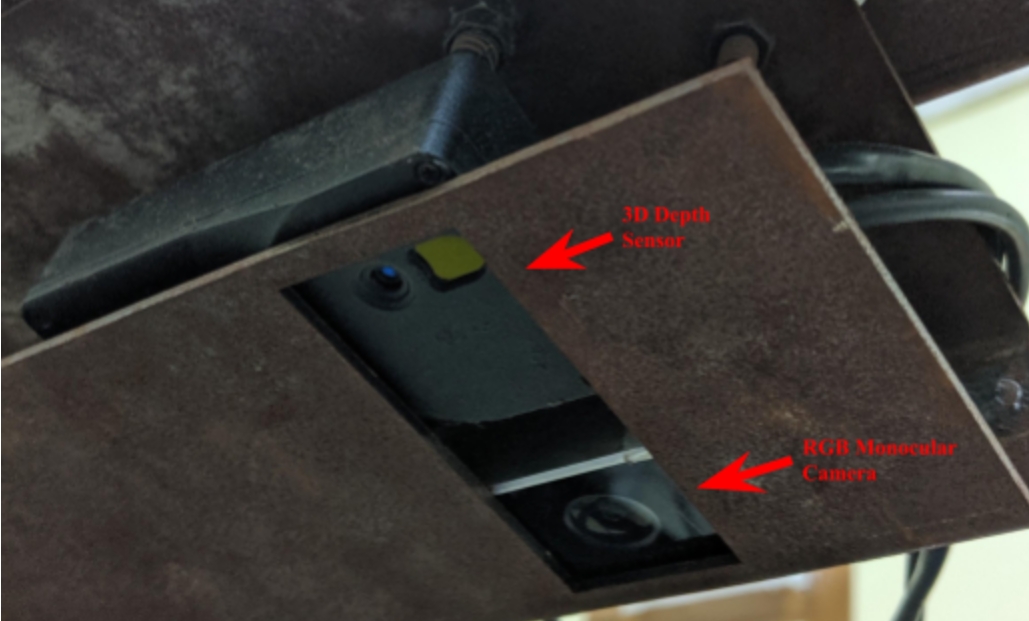}
\captionof{figure}{Dual Vision System consisting of RGB Camera and 3D Depth Sensor}
\end{figure}

\noindent The dual sensor setup shown in Figure 2 consists of a monocular camera and the 3D depth sensor placed parallel to each other on a rigid mount. The segmentation of the box is done using the RGB information from the monocular camera. A binary mask is generated for the segmented boxes, where the detected boxes have a value of one and rest as zero. Further for the pixels in the mask, corresponding points in the point cloud are considered for fitting a plane to obtain its centroid and orientation. Finally, for sending obtained 6DOF information of the box detected by the dual vision system to the robot, a calibration procedure \cite{OpenCV} is carried out between the ABB IRB 1200 robot and the 3D depth sensor to transform the coordinates obtained with respect to the 3D depth sensor frame to the robot base frame.

\begin{figure}[h]
\centering
\includegraphics[scale=0.5]{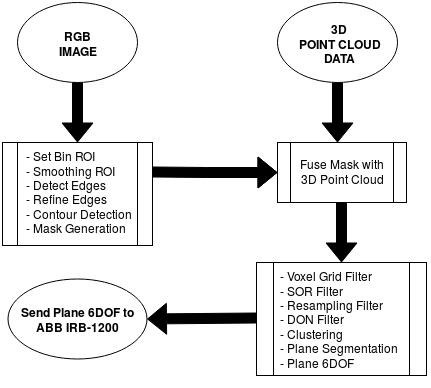}
\captionof{figure}{Dual Vision System consisting of RGB Camera and 3D Depth Sensor}
\end{figure}


\subsection{Box Segmentation}
\noindent To segment out boxes, we have incorporated a robust edge detection algorithm, which detects all possible edges in the image and filters out the unnecessary edges, while preserving the structural properties of the detected boxes to be used for further processing. The proposed algorithm also performs efficiently for differentiating two objects kept very close to each other. The algorithm further consists of a parent-child hierarchy, such that if a small box is placed on top of a large box, the identity of the smaller box should be preserved within the larger box. The smaller box is termed as a child and the larger box is termed as a parent by the algorithm. The child box is given a priority to be picked first by the algorithm.

\begin{figure}[t]
	\begin{subfigure}{.15\textwidth}
	\centering
	\includegraphics[width=1.0\linewidth]{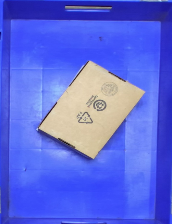}
	\caption{ROI of bin}
	\end{subfigure}
	\begin{subfigure}{.15\textwidth}
	\centering
	\includegraphics[width=1.0\linewidth]{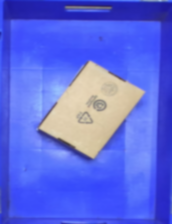}
	\caption{Smoothening}
	\end{subfigure}
	\begin{subfigure}{.15\textwidth}
	\centering
	\includegraphics[width=1.0\linewidth]{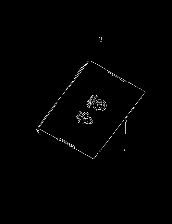}
	\caption{Gradients}
	\end{subfigure}
	
	\begin{subfigure}{.15\textwidth}
	\centering
	\includegraphics[width=1.0\linewidth]{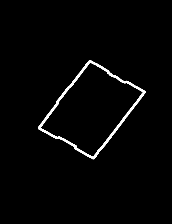}
	
	\caption{Refining Gradient}
	\end{subfigure}
	\begin{subfigure}{.15\textwidth}
	\centering
	\includegraphics[width=1.0\linewidth]{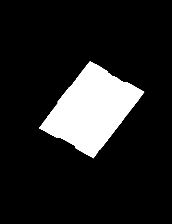}
	\caption{Mask Generation}
	\end{subfigure}

	\caption{Results of box segmentation algorithm on individual object}
\end{figure}

The algorithm runs in 6 separate steps:

\textbf{Set Bin ROI}: In the image taken from the camera, the region of interest(ROI) of the bin is extracted and used in the further processing of the algorithm. Bin ROI is shown in Figure 4a.

\textbf{Smoothing}: Images taken from a camera might contain some amount of noise. To prevent that noise from being mistaken for edges, a gaussian filter with kernel 3x3 is applied to smoothen the input image as shown in Figure 4b.

\textbf{Finding Gradients}: Popular Canny edge algorithm \cite{Canny} is known for finding edges whose intensity gradients fall in a range of predefined threshold values. We have incorporated an auto canny algorithm \cite{AutoCanny} that adaptively decides the threshold based on the region of interest of the bin in the image. The edges are determined by applying gradient at each pixel in the smoothened image using Sobel operator as shown in Figure 4c. To approximate the gradient in the x- and y-direction respectively the kernels shown in Equation \ref{eqn_SobelKernel1} and \ref{eqn_SobelKernel2} are used.
\begin{equation}
\centering
KGX=
	\begin{bmatrix}
		\hspace{0.1cm}-1 & \hspace{0.2cm}0 & \hspace{0.2cm}1 \hspace{0.15cm} \\
		\hspace{0.1cm}-2 & \hspace{0.2cm}0 & \hspace{0.2cm}2 \hspace{0.15cm}\\
		\hspace{0.1cm}-1 & \hspace{0.2cm}0 & \hspace{0.2cm}1 \hspace{0.15cm}\\
	\end{bmatrix}
\label{eqn_SobelKernel1}
\end{equation}

\begin{equation}
KGY=
	\begin{bmatrix}
		\hspace{0.2cm}1 & \hspace{0.2cm}2 & \hspace{0.2cm}1 \hspace{0.15cm} \\
         \hspace{0.2cm}0 & \hspace{0.2cm}0 & \hspace{0.2cm}0 \hspace{0.15cm} \\
       -1 & -2 & -1 \hspace{0.15cm} \\     
	\end{bmatrix}  
\label{eqn_SobelKernel2}
\end{equation}

\textbf{Refining Gradients}: Over the edges obtained in the image from the auto canny edge detection, contours are detected. These are further refined over the contour area of 2500 pixel-sq in a 2K pixel density image to remove unnecessary edges. Contour detection aims at quantifying the presence of a closed boundary at a given image location through local measurements. This step results in the identification of surfaces of boxes as shown in Figure 4d.

\textbf{Parent-Child hierarchy}: A contour hierarchy is implemented to solve the stacking of smaller boxes over larger boxes. Contour detection is used to detect objects in the image, but in some cases, objects are present on top of another object, just like nested figures. In this case, we call the outer one as a parent and the inner one as a child. This way, contours in an image constitute a relationship with each other. Representation of this relationship is called the Hierarchy. An individual object is determined as a parent contour as shown in Figure 4d.

\textbf{Mask Generation}: A binary mask is generated for the child and parent boxes individually, where the full surface of the detected box has a value of one and rest part of the bin as zero. While the child contour is given a priority, the parent contour has a zero value. The mask is generated for parent contours only after all the child contours have been picked by the ABB 1200 robot as shown in Figure 4e. This binary mask is further used on the point cloud to detect planes and generate the 6DOF of the plane.

\subsection{Box Localization}
This section discusses plane fitting and 6 DOF pose estimation of the object. The masks calculated in the previous section are fused with the raw 3D point cloud obtained from the depth sensor. The results will depend on the vision system calibration step performed \cite{OpenCV}. Finally, we have a cloud scene with separated planes for all the objects as shown in Figure 6a. As this is raw from the sensor, it tends to be a bit noisy. We will further process this data in five steps to finally segment out planes and estimate centroid and Euler angles for the robot to grasp. These steps are described in the following subsections.

\SetInd{0.5em}{0.2em}
\begin{algorithm}[h!]
\small
 \KwIn{Masked Point Cloud Clusters (clusters)}
 \KwOut{Planes, Model Coefficients, Plane Normals}

 $uni\_plane,uni\_plane\_coeff,normal\_vecs = []$\
 
 \For {$cluster \in clusters$}{
 	$prev\_coeff, planes, plane\_coeff = []$\
 	
 	\While{$cluster.size() \geq min\_cluster\_size$}{
 		$points,coeff = ransac\_seg\_plane(cluster)$
 		
 		\If{$prev\_coeff == coeff$}{
 			$break$
 		}
 		\If{$points.size() \geq min\_object\_size$}{
 			$plane.append(points)$
 			
 			$plane\_coeff.append(coeff)$
 			
 			$cluster = rm\_seg\_plane(cluster, points)$\
 		}
		$plane\_num = planes.size()$
		
		\If{$plane\_num == 0$}{
			continue
		}
		\Else{
			$dp\_mat = dot\_product(normal\_vecs)$
			
			$group, visited = []$
			
			\For{$m \in range(plane\_num)$}{
				\If{$m \not \in visited$}{
					$visited.append(m)$ 
					
					$g = []$
					
					$g.append(m)$
					
					\For{$n <= size(plane\_num)$}{
						\If{$n \not \in visited\ \&\ dp\_mat[m][n] \sim 1.0$}{
							\If{$check\_overlapping(planes[m], planes[n])$}{
								$g.append(n)$
							}
						}	
					}
				}
			}
			$group.append(g)$
		}
		\For{$g \in group$}{
			$uni\_plane.append(comp\_grp\_plane(g))$
			
			$uni\_plane\_coeff.append(comp\_grp\_plane\_coeff(g))$
			
			$normal\_vecs.append(comp\_grp\_plane\_normal(g))$
		}		
 	}
 }
  
\Return{$uni\_plane,uni\_plane\_coeff,normal\_vecs$}
\newline
\caption{Plane Segmentation and Extraction}
\label{algo_boxlocalization}
\end{algorithm}

\subsubsection{Filtering}

Since the data consists of thousands of points, it will take a huge amount of time to directly compute planes on this data. Moreover, the data is prone to noise points because of sensors' measurement capability and repeatability. There are various methods available for filtering the point cloud \cite{Xian-Feng-et-al}, \cite{Moreno and Ming L.}. We have selected two filters based on their efficiency and time complexity to reduce the number of points and to minimize noise, after which an output shown in Figure 6b is obtained.

Firstly, we apply the Voxel Grid filter to downsample the cloud. This method assumes a 3D voxel grid over the cloud and then in for each voxel, a centroid is assumed to approximate all the points in that voxel. This method robustly reduces the number of points based on the leaf size which defines the size of voxel and it represents the underlying surface more accurately. We can optimally set this parameter so that geometric features don’t get lost and the number of points also gets significantly reduced.     

Secondly, we use Statistical Outlier Removal (SOR) filter \cite{HarisBalta-et-al}. Generally, point clouds registered from depth sensors have some sparse outliers which can corrupt the estimation of characteristics like surface normals. SOR computes a distribution of the mean distance of a point to all its k nearest neighbors specified by a parameter. The resulting distribution is assumed to be Gaussian with a mean and a standard deviation. All the points whose mean distances are above the threshold defined by the global distances mean and standard deviation parameters are classified as outliers and removed from the cloud.

\subsubsection{Resampling and Difference of Normals} 
Filtering, discussed in the above section, is incapable of removing some erroneous points caused by measurement errors. Thus we perform resampling using Moving Least Squares (MLS) surface reconstruction method \cite{Fleishman-et-al}. This resampling technique also helps in accounting irregularities in the cloud created due to glossy surfaces and occlusions. The algorithm estimates missing points in the surfaces by using higher-order polynomial interpolations between the neighboring data points. 

Following this, we perform a Difference of Normals based segmentation \cite{Ioannou-et-al} to remove the edges between the planes of the cuboidal objects. The normals calculated on the plane point cloud are shown in Figure 6c. This method enhances the performance of plane segmentation algorithm discussed later by separating the faces of boxes. We first calculate two unit point normals \textbf{\^{n}}(\textbf{p},\textit{r$_{l}$}), \textbf{\^{n}}(\textbf{p},\textit{r$_{s}$}) for every point \textit{p} such that \textit{r$_{l}$} $>$ \textit{r$_{s}$} as shown in Figure 5. Succeeding this we calculated a normalized Difference of Normals vector field using equation \ref{eqn_don}, where r$_{s}$, r$_{l}$ $\in \mathbb{R}$, r$_{s}$ $<$ r$_{l}$.
 
\begin{equation}
\Delta\hat{n}(p, r_{s}, r_{l})=\frac{\hat{n}(p, r_{s}) - \hat{n}(p, r_{l})}{2}
\label{eqn_don}
\end{equation}

Taking the norm of this vector field yields values in the range (0,1) for all the points. We select the points for which the norm has value less than a user-defined threshold parameter.

\begin{figure}[h]
    \centering
	\includegraphics[scale=0.24]{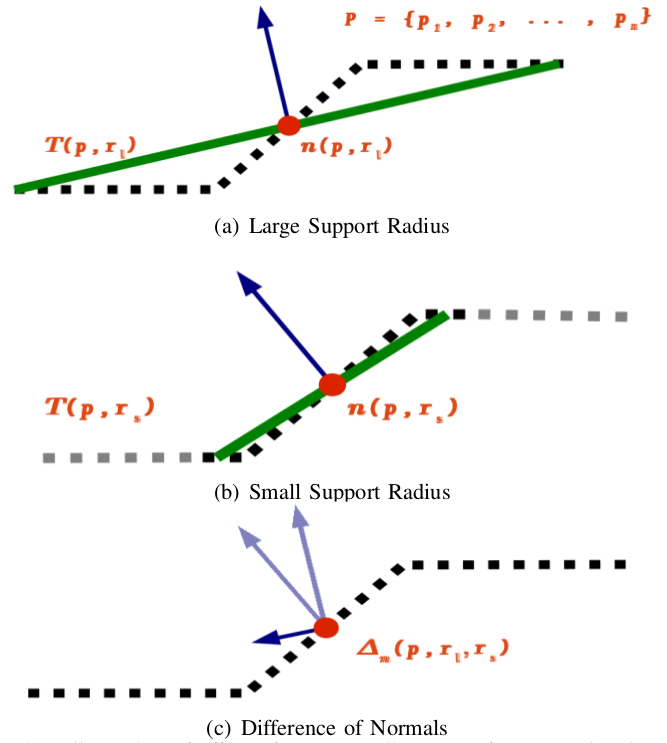}
	\captionof{figure}{DoN : Illustration of effect of support radius on surface normal estimates in 1D}
	\label{fig_don}
\end{figure}

\subsubsection{Clustering} 
Clustering is an important step as it helps ensure that a common plane is not fitted for two parallel faced cuboidal objects. Our goal here is to accurately generate separate clusters for all the objects. An algorithm is required which is robust to varying density in the cloud and selects clusters based on hierarchy. Thus we use HDBSCAN \cite{Campello-et-al} algorithm which predicts stable clusters in noisy environments and is not prone to outliers. The algorithm works by transforming the cloud-based on sparsity and then constructs a minimum spanning the tree of the distance weighted graph. Following this, cluster hierarchy is generated for connected components in the graph. Finally, stable clusters are selected by condensing the cluster hierarchy based on the minimum cluster size. This minimum cluster size is a user-defined parameter and should be set equal to the number of points on the surface of the smallest object. For our application, we have set a minimum cluster size equal to 30 points and obtained the output for a plane as shown in Figure 6d.

\subsubsection{Plane Segmentation} 
Plane Segmentation is the most critical part of the algorithm. The main challenge here is to accurately segment unique planes such that no two planes belong to the same surface. The clusters generated in the previous section may contain more than one plane. Thus plane fitting is done iteratively to segment out all the planes in that cluster \cite{Oehler-et-al}.  We use the Random Sample Consensus (RANSAC) algorithm \cite{Fischler and Bolles} with a plane model over the cloud clusters. 

RANSAC algorithm starts by selecting a random subset of points from the cluster, called hypothetical inliers. Model is fitted for these hypothetical inliers and all the remaining points are checked against this model to be classified as inliers or outliers. Then this model is again re-estimated by using all classified inliers. This process is repeated until we get a segmented plane with its coefficients from the cluster. The coefficients of this segmented plane are saved and all the points belonging to this plane are removed from the cluster. The updated cluster is again processed for plane segmentation. This process keeps on repeating until the number of points in the cluster becomes less than the user-defined parameter for minimum points in the cluster. A similar kind of processing is done for all the clusters. The parameters of the RANSAC algorithm should be chosen wisely such that the points which are not on the surface to be grasped don't become part of the segmented plane.

Following the segmentation of all the planes from all the clusters, we need to extract unique planes. Generally, due to measurement noise in sensors, the same surface can be detected as 2 different planes. In order to address this problem, we group all segmented planes which are parallel to each other. The two parallel planes found from different clusters (Figure 6d) after RANSAC but are part of the same surface are combined into a single plane as shown in Figure 6e. This is done by combining all those planes which are overlapping using the following checks:

\begin{itemize}
\item Distance between the centroid is less than a certain threshold. This threshold is defined based on the minimum object size to be picked. 
\item Perpendicular distance between planes is less than the measurement noise of the depth sensor. Correct selection of this parameter ensures that planes which are part of same surface gets combined.
\end{itemize}

The pseudo code shown in \ref{algo_boxlocalization} contains the functions \emph{check\_overlapping} to check for overlapping planes and \emph{comp\_grp\_plane} to group overlapping planes.

\subsubsection{6DOF Pose Estimation} 
After finding all the unique planes from the cloud, we need to generate a 6DOF pose over each plane for the robot to grasp the object as shown in Figure 6f. Since in our implementation we used a vacuum gripper, we need to calculate the centroid of the plane and Euler angles to represent the orientation of the plane.

The centroid of the plane is calculated using Mean Shift clustering \cite{Comaniciu and Meer} method. The advantage of using this algorithm over others is that it is a centroid based algorithm. This algorithm almost accurately calculates the center of the region even in the case of planes having sparse points. The centroid is calculated with reference to the depth sensor frame.

We estimated Euler angles by attaching a reference frame at the previously computed centroid for each plane. For each plane, we have a normal vector from estimated plane coefficients from which we approximate an axis along the plane by finding a unit vector along the plane. Finally, we compute the cross product between the unit normal vector and unit vector along the plane to find the third axis of the reference frame. Following this, a rotation matrix is computed for defining the rotation of this object frame with respect to the depth sensor. Further, this rotation matrix is converted into Z-Y-X Euler angles with respect to the depth sensor frame.

\begin{equation}
	Z_1Y_2X_3 = 
	\begin{bmatrix}
		c_1c_2 & c_1s_2s_3 - c_3s_1 & s_1s_3 + c_1c_3s_2 \\
		c_2s_1 & c_1c_3 + s_1s_2s_3 & c_3s_1s_2 - c_1s_3 \\
		-s_2 & c_2s_3 & c_2c_3 \\ 
	\end{bmatrix}
	\label{eqn_euler}
\end{equation}
,where c represents cos(), s represents sin(), 1,2 and 3 are Euler angles about X, Y and Z axes respectively.

\begin{figure}[t]
	\begin{subfigure}{.15\textwidth}
	\centering
	\includegraphics[width=1.0\linewidth]{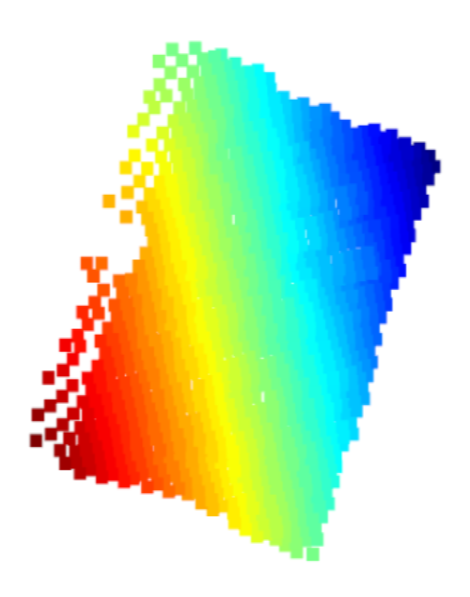}
	\caption*{\textbf{(a)}Mask Fused Cloud}
	\end{subfigure}
	\begin{subfigure}{.15\textwidth}
	\centering
	\includegraphics[width=1.0\linewidth]{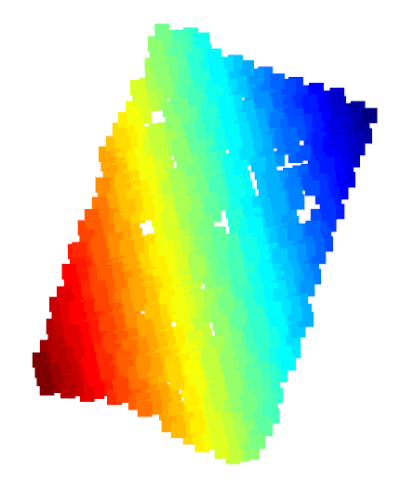}
	\caption*{\textbf{(b)}Filtering}
	\end{subfigure}
	\begin{subfigure}{.15\textwidth}
	\centering
	\includegraphics[width=1.0\linewidth]{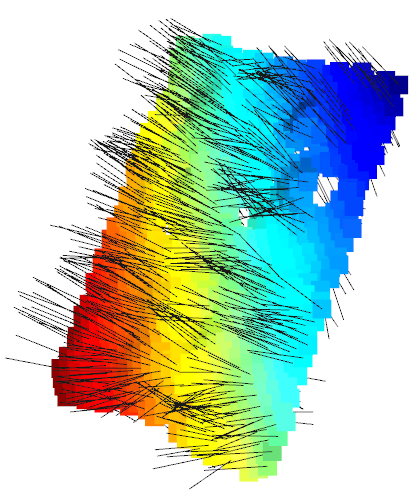}
	\caption*{\textbf{(c)}Re-sampling and DON}
	\end{subfigure}
	\begin{subfigure}{.15\textwidth}
	\centering
	\includegraphics[width=1.0\linewidth]{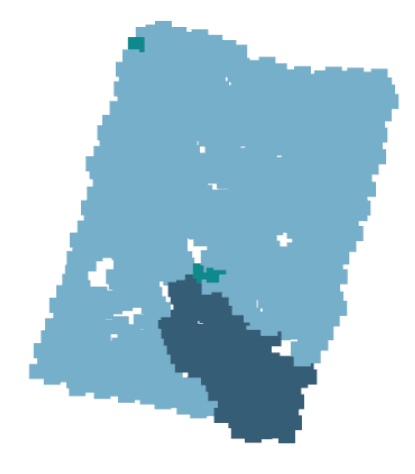}
	\caption*{\textbf{(d)}Clustering}
	\end{subfigure}
	\hspace{0.1cm}
	\begin{subfigure}{.15\textwidth}
	\centering
	\includegraphics[width=1.0\linewidth]{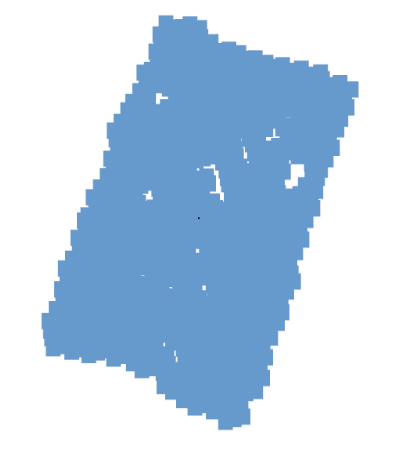}
	\caption*{\textbf{(e)}Plane Segmentation}
	\end{subfigure}
	\hspace{0.1cm}
	\begin{subfigure}{.15\textwidth}
	\centering
	\includegraphics[width=1.0\linewidth]{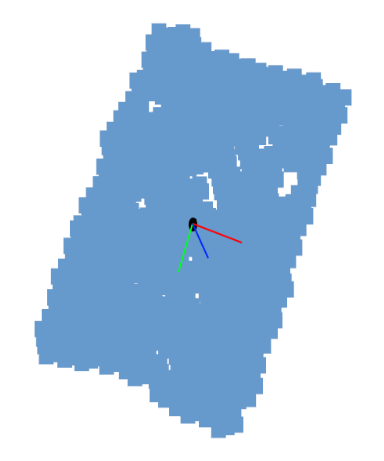}
	\caption*{\textbf{(f)}Pose Estimation}
	\end{subfigure}

	\caption{Results of Box Localisation algorithm for individual object}
\end{figure}


\section{Experiment Test-bench}
We performed our experiments with ABB IRB 1200 robot and dual vision system. The dual vision system consisted of a Monocular RGB camera with 2K resolution and a Time of Flight based 3D depth sensor which gives an organized point cloud, of resolution 224x172. This visual system was mounted at a height of 1.2m above the bin having dimensions 400mm x 600mm. All the experiments were performed on a laptop having Ubuntu 16.04 with a 2.30GHz Intel Core i5-8300H CPU and 4GB DDR3 RAM.

\section{Results}
This section is divided into three subsections: 
\subsection{Verification}
To evaluate the accuracy of the 3D pose estimated by our system, a plate with an Aruco marker \cite{Kam-et-al} is attached to the last link of the robot with a known static transform as shown in Figure 7. 

\begin{figure}[h]
    \centering
	\includegraphics[scale=0.03]{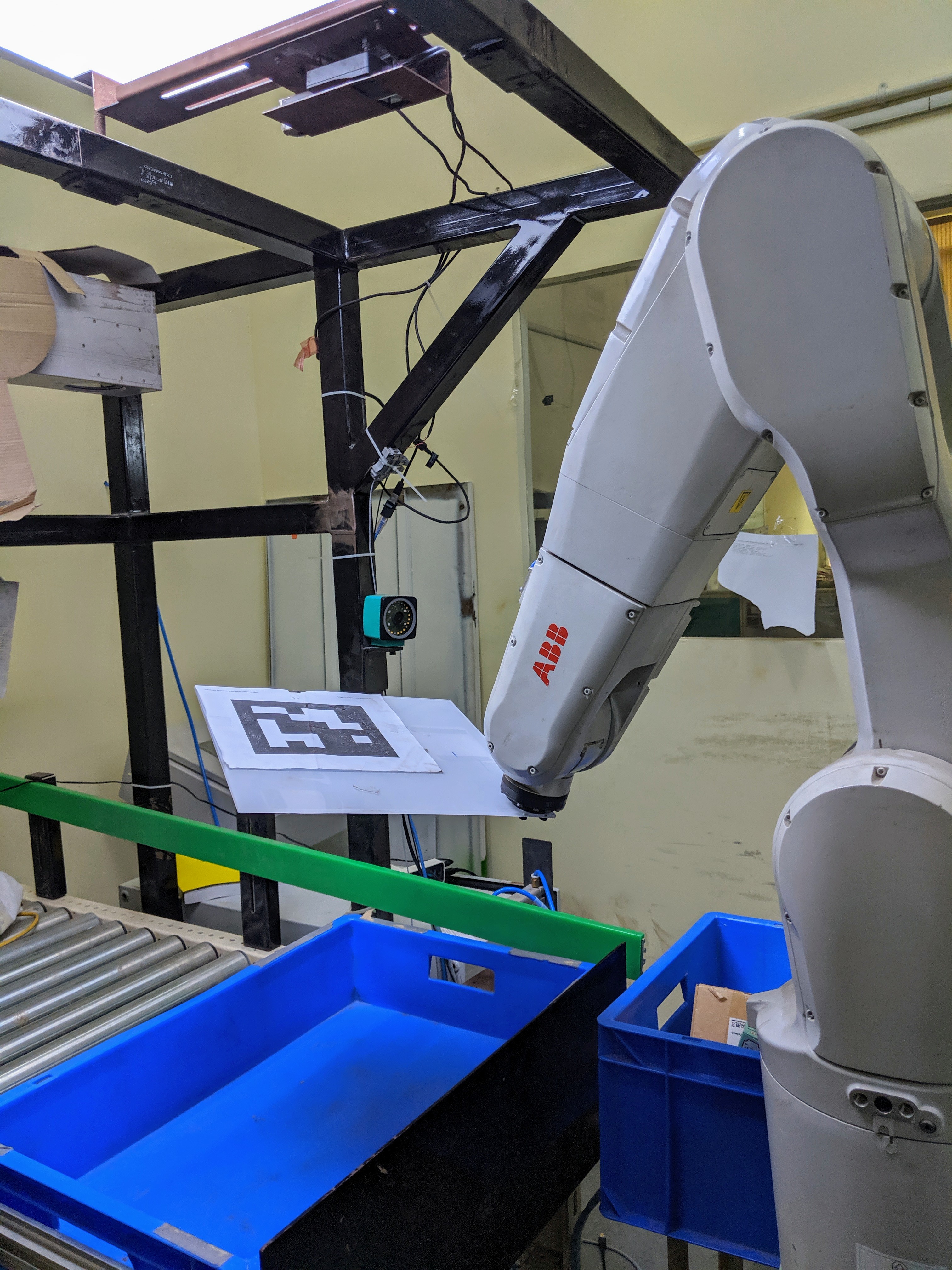}
	\captionof{figure}{Experimental setup for performing verification procedure}
	\label{fig_don}
\end{figure}

\noindent The robot is moved by the teach pendant such that the plate gets all possible orientations of a box in the bin. The Aruco maker is detected in the RGB image of the camera and a mask is generated for it \cite{Kam-et-al}. Further, a plane is fitted in the point cloud for this mask and its 6DOF pose is estimated by our system which is compared with the ground truth values of the teach pendant. In our experiment, the performance was tested on a total of  9 orientation cases of the plate, For each test case, we have recorded 20 samples and taken the average of the error in pose calculated by the algorithm. All the Euler angles are measured with respect to the robot base frame and the z-axis of the robot is pointing upward with respect to the ground. The results obtained for the 9 test cases of the plate are shown in Table 1.  Additionally, the system demonstrated stable performance with an overall mean translation error of less than 3.3 mm in all directions and a mean rotation error of less than 3.26 $\deg$ on all axes.

\begin{table}[h]
\centering
\begin{adjustbox}{width=0.48\textwidth, height=0.1\textwidth}
\begin{tabular}{|c|c|c|c|c|c|c|c|}
\hline
\textbf{Description} & \textbf{Angle} &  \textbf{Angle} & \textbf{Trans Err} & \textbf{Trans Err} & \textbf{Trans Err} &  \textbf{Rot Err} & \textbf{Rot Err} \\
             & \textbf{X}     & \textbf{Y}      & \textbf{X(in mm)} & \textbf{Y(in mm)} & \textbf{Z(in mm)} & \textbf{X(deg)}  & \textbf{Y(deg)} \\
\hline
+ve X                   & 135                  & 0   & 2.7                                    & 2.3          & 3.4          & 2.8                                    & 2.9               \\
\hline
-ve X                   & -135                 & 0   & 2.4                                    & 2.5          & 3.1          & 2.9                                    & 2.6               \\
\hline
+ve Y                   & 180                  & 45  & 3.1                                    & 2.7          & 2.9          & 2.7                                    & 3.1               \\
\hline
-ve Y                   & 180                  & -45 & 2.9                                    & 2.9          & 2.8          & 2.8                                    & 3.2               \\
\hline
+ve X and +ve Y             & 135                  & 45  & 4.7                                    & 4.8          & 3.7          & 3.5                                    & 4.1               \\
\hline
-ve X and +ve Y & -135                 & 45  & 2.7                                    & 3.1          & 3.4          & 3.1                                    & 3.8               \\
\hline
+ve X and -ve Y  & 135                  & -45 & 2.6                                    & 3.2          & 3.7          & 3.3                                    & 3.4               \\
\hline
-ve X and +ve Y             & -135                 & -45 & 4.6                                    & 4.7          & 3.5          & 3.4                                    & 3.9               \\
\hline
Parallel to bin surface                     & 180                  & 0   & 1.6                                    & 2.1          & 3.2          & 2.1                                    & 2.4               \\
\hline
 \textbf{Average Error}                                            &                      &     & \textbf{3.03}                                   & \textbf{3.27}         & \textbf{3.3}          & \textbf{2.95}                                   & \textbf{3.26} \\
\hline            
\end{tabular}
\end{adjustbox}
\caption{Computation of system error for 6DoF pose.} 
\end{table}

\begin{table}[h]
\centering
\begin{adjustbox}{width=0.48\textwidth, height=0.05\textwidth}
\tiny
\begin{tabular}{|c|c|}
\hline
\textbf{Box Surface Type} & \textbf{Mean Trans Error from box centre (in mm)} \\
\hline
No texture       & 2.1                                         \\
\hline
Minimal Texture  & 2.8                                         \\
\hline
Highly textured  & 3.6                                         \\
\hline
Glossy Surface   & 5.7										\\
\hline                                        
\end{tabular}
\end{adjustbox}
\caption{Average error for different box surface textures measured over 25 samples} 

\end{table}

\begin{figure}[b!]
    \centering
	\begin{subfigure}{.09\textwidth}
	\centering
	\includegraphics[width=.9\linewidth]{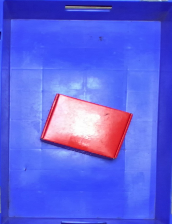}
	\end{subfigure}
	\begin{subfigure}{.09\textwidth}
	\centering
	\includegraphics[width=.9\linewidth]{final_images/minimal_texture/image3.png}
	\end{subfigure}
	\begin{subfigure}{.09\textwidth}
	\centering
	\includegraphics[width=.9\linewidth]{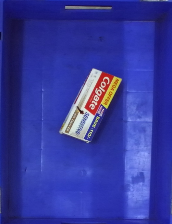}
	\end{subfigure}
	\begin{subfigure}{.09\textwidth}
	\centering
	\includegraphics[width=.9\linewidth]{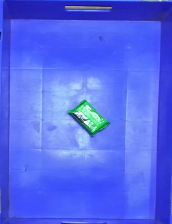}
	\end{subfigure}
	
	\begin{subfigure}{.09\textwidth}
	\centering
	\includegraphics[width=.9\linewidth]{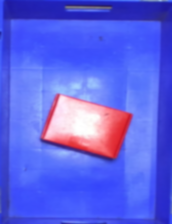}
	\end{subfigure}
	\begin{subfigure}{.09\textwidth}
	\centering
	\includegraphics[width=.9\linewidth]{final_images/minimal_texture/blurred3.png}
	\end{subfigure}
	\begin{subfigure}{.09\textwidth}
	\centering
	\includegraphics[width=.9\linewidth]{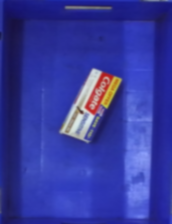}
	\end{subfigure}
	\begin{subfigure}{.09\textwidth}
	\centering
	\includegraphics[width=.9\linewidth]{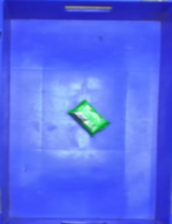}
	\end{subfigure}
		
	\begin{subfigure}{.09\textwidth}
	\centering
	\includegraphics[width=.9\linewidth]{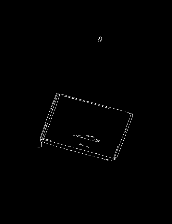}
	\end{subfigure}
	\begin{subfigure}{.09\textwidth}
	\centering
	\includegraphics[width=.9\linewidth]{final_images/minimal_texture/edges3.png}
	\end{subfigure}
	\begin{subfigure}{.09\textwidth}
	\centering
	\includegraphics[width=.9\linewidth]{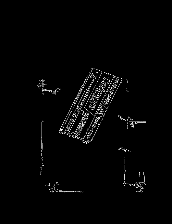}
	\end{subfigure}
	\begin{subfigure}{.09\textwidth}
	\centering
	\includegraphics[width=.9\linewidth]{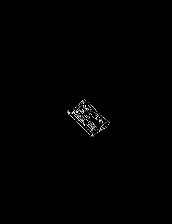}
	\end{subfigure}
	
	\begin{subfigure}{.09\textwidth}
	\centering
	\includegraphics[width=.9\linewidth]{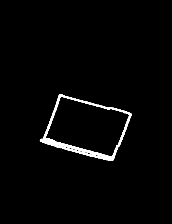}
	\end{subfigure}
	\begin{subfigure}{.09\textwidth}
	\centering
	\includegraphics[width=.9\linewidth]{final_images/minimal_texture/refined3.png}
	\end{subfigure}
	\begin{subfigure}{.09\textwidth}
	\centering
	\includegraphics[width=.9\linewidth]{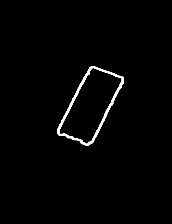}
	\end{subfigure}
	\begin{subfigure}{.09\textwidth}
	\centering
	\includegraphics[width=.9\linewidth]{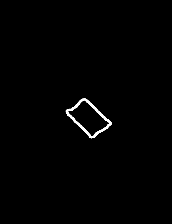}
	\end{subfigure}
	
	\begin{subfigure}{.09\textwidth}
	\centering
	\includegraphics[width=0.9\linewidth]{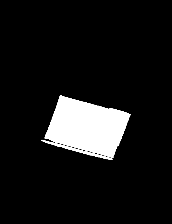}
	\caption{No Texture}
	\end{subfigure}
	\begin{subfigure}{.09\textwidth}
	\centering
	\includegraphics[width=.9\linewidth]{final_images/minimal_texture/mask3.png}
	\caption{Min Texture}
	\end{subfigure}
	\begin{subfigure}{.09\textwidth}
	\centering
	\includegraphics[width=.9\linewidth]{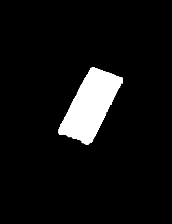}
	\caption{High Texture }
	\end{subfigure}
	\begin{subfigure}{.09\textwidth}
	\centering
	\includegraphics[width=.9\linewidth]{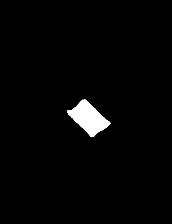}
	\caption{Glossy Surface}
	\end{subfigure}
	\caption{Segmentation results on boxes with different surface types as mentioned in test cases}
\end{figure}

\subsection{Box Test Case}
The surface details of the boxes used in experimentation are presented in Table 2. We have used boxes with a minimum dimension of 50mmX75mmX20mm and various boxes having a surface with no texture, minimal texture, maximum texture, and gloss. Further, some special test cases are taken into consideration where two objects of the same dimensions are kept adjacent to each other and a smaller object is kept on top of a larger object. For planar surfaces, the deviation along the x and y axes determines the accuracy of the system. This accuracy further determines if the system is good enough to perform random boxes picking tasks, whereas the deviation and orientation on the z-axis with respect to the gripping tool are less important as the single suction cup type gripper can easily search for the box within a certain range in the z-direction. After the box being picked by the robot, it is put in the barcode scanning zone where it is identified by the system. The maximum average percentage error in the result of the translation estimate is found to be 3.12\% along the x and y axes of the box.  This error is acceptable as the box is not to be picked from its barcode which is generally present at the edges of the box. The error computed ensured that the proposed method worked well on the boxes in the cluttered environment for random bin-picking tasks.

\subsection{Complete System Performance}
The performance of our system for the box specifications mentioned in Table 2 is shown in Figure 8. The boxes are placed individually in the bin and the image from the RGB camera is first processed using the box segmentation module and then fused with the point cloud data of the depth sensor. This is given as input to the box localization module to extract the 6DOF pose data of the box.

Figure 8a shows the performance for the box with no texture, Figure 8b shows performance for the box with minimal texture, Figure 8c shows performance for the box with high texture and Figure 8d shows performance for the box with a glossy surface. The pose estimation error is slightly higher in the case of glossy surfaces because of sensor limitations caused due to reflectivity.

\begin{figure}[t]
    \centering
	\begin{subfigure}{.1\textwidth}
	\centering
	\includegraphics[width=.9\linewidth]{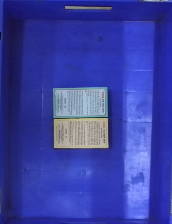}
	\end{subfigure}
	\hspace{1.5cm}
	\begin{subfigure}{.1\textwidth}
	\centering
	\includegraphics[width=.9\linewidth]{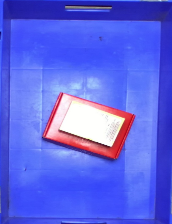}
	\end{subfigure}
		
	\begin{subfigure}{.1\textwidth}
	\centering
	\includegraphics[width=.9\linewidth]{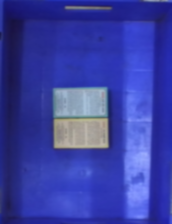}
	\end{subfigure}
	\hspace{1.5cm}
	\begin{subfigure}{.1\textwidth}
	\centering
	\includegraphics[width=.9\linewidth]{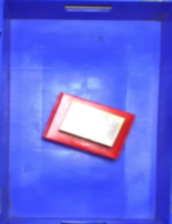}
	\end{subfigure}
	
	\begin{subfigure}{.1\textwidth}
	\centering
	\includegraphics[width=.9\linewidth]{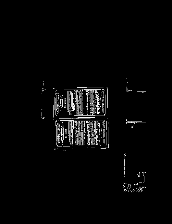}
	\end{subfigure}
	\hspace{1.5cm}
	\begin{subfigure}{.1\textwidth}
	\centering
	\includegraphics[width=.9\linewidth]{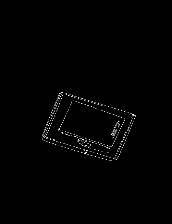}
	\end{subfigure}

	\begin{subfigure}{.1\textwidth}
	\centering
	\includegraphics[width=.9\linewidth]{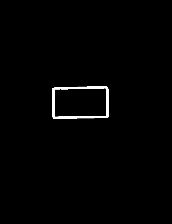}
	\end{subfigure}
	\hspace{1.5cm}
	\begin{subfigure}{.1\textwidth}
	\centering
	\includegraphics[width=.9\linewidth]{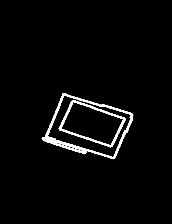}
	\end{subfigure}
			
	\begin{subfigure}{.1\textwidth}
	\centering
	\includegraphics[width=0.9\linewidth]{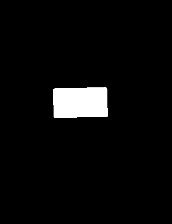}
	\end{subfigure}
	\hspace{1.5cm}
	\begin{subfigure}{.1\textwidth}
	\centering
	\includegraphics[width=.9\linewidth]{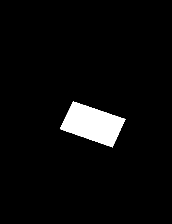}
	\end{subfigure}
	
	\begin{subfigure}{.1\textwidth}
	\centering
	\includegraphics[width=.9\linewidth]{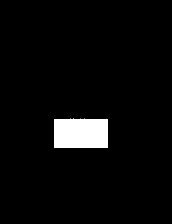}
	\caption*{\textbf{(a)}Adjacent Boxes}
	\end{subfigure}
	\hspace{1.5cm}
	\begin{subfigure}{.1\textwidth}
	\centering
	\includegraphics[width=.9\linewidth]{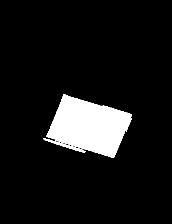}
	\caption*{\textbf{(b)}Overlapping Boxes}
	\end{subfigure}
	\caption{Results of mask generation for Parent Child Hierarchy for Adjacent and Overlapped test case}

\end{figure}

The performance of our system in the case of two adjacent boxes of the same dimensions is shown in Figure 9a and the case of smaller box placed on top of a larger box is shown in Figure 9b. Where the 1st row shows the input RGB image ROI of the bin, 2nd row shows the smoothened image, 3rd row shows the Edges found, 4th row shows the refined edges found, 5th row shows the mask of child contour and the 6th row shows the mask of parent contour, after the child was picked by the robot.

\begin{figure}[h!]
    \centering
	\begin{subfigure}{.09\textwidth}
	\centering
	\includegraphics[width=.9\linewidth]{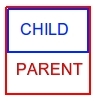}
	\end{subfigure}
	\hspace{1.5cm}
	\begin{subfigure}{.09\textwidth}
	\centering
	\includegraphics[width=.9\linewidth]{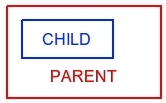}
	\end{subfigure}
	
	\begin{subfigure}{.09\textwidth}
	\centering
	\includegraphics[width=.9\linewidth]{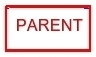}
	\caption*{\textbf{(a)}Adjacent boxes}
	\end{subfigure}
	\hspace{1.5cm}
	\begin{subfigure}{.09\textwidth}
	\centering
	\includegraphics[width=.9\linewidth]{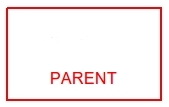}
	\caption*{\textbf{(b)}Overlapped boxes}
	\end{subfigure}
	\caption{Parent Child Hierarchy}
\end{figure}

\noindent  Figure 10 explains the parent-child hierarchy in both cases. When two adjacent boxes are taken into consideration as shown in Figure 10a, they are not identified as two individual boxes. Instead, their contours are detected as one contour (child) is inside another contour(parent) and the child is given a priority to be picked up. Once the child box is picked up, in the next detection the parent box is picked up. When the case of a smaller box is kept on top of a larger box is taken into consideration as shown in Figure 10b, the smaller box is detected as child contour and the larger box is detected as parent contour. The vision system first segments the child contour boxes and computes their midpoint [x, y, and z coordinates] and orientation [Euler angles]  and further goes for the parent contour boxes when it has picked all the child contour boxes. Figure 11 shows the performance of our system in the cluttered bin. Our proposed algorithm was able to detect smaller boxes on the top of larger boxes due to the parent-child hierarchy used in generating masks. These masks were further used to extract an accurate plane from the point cloud and estimate the 6DoF pose. 

\begin{figure}[h]
    \centering
	\begin{subfigure}{.15\textwidth}
	\centering
	\includegraphics[width=1.0\linewidth]{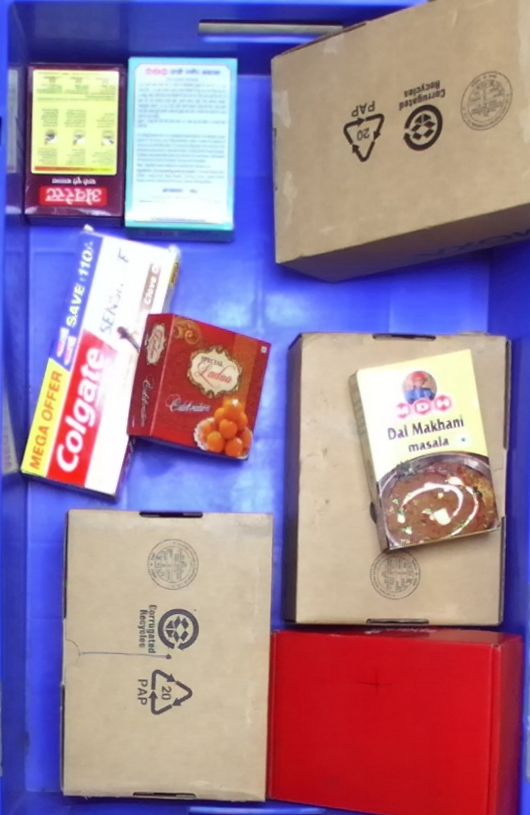}
	\caption{Bin ROI}
	\end{subfigure}
	\begin{subfigure}{.15\textwidth}
	\centering
	\includegraphics[width=1.0\linewidth]{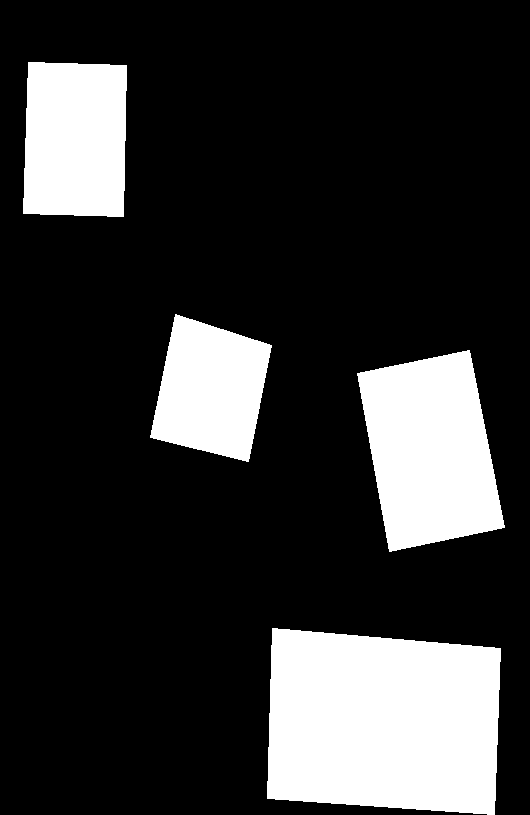}
	\caption{Child Mask}
	\end{subfigure}
	\begin{subfigure}{.15\textwidth}
	\centering
	\includegraphics[width=1.0\linewidth]{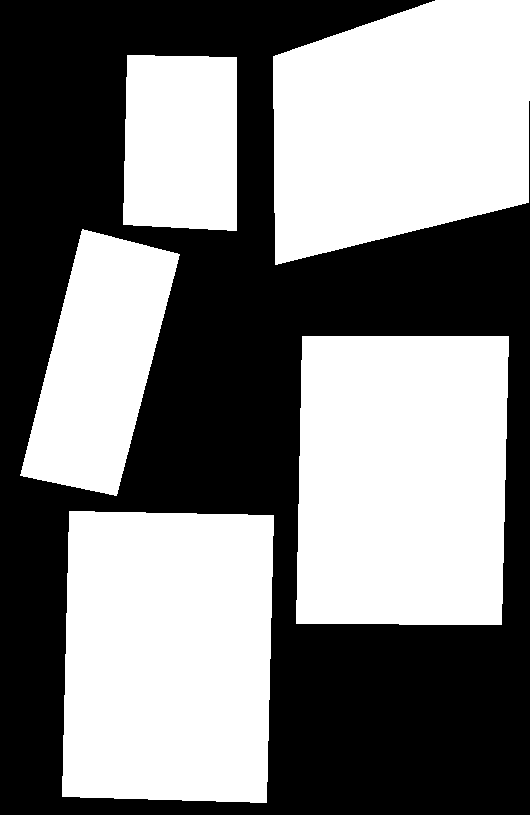}
	\caption{Parent Mask}
	\end{subfigure}
	
	\caption{Results of algorithm for Cluttered environment case}
\end{figure}

\begin{table}[t]
\centering
\begin{adjustbox}{width=0.48\textwidth, height=0.08\textwidth}
\begin{tabular}{|c|c|}
\hline
\textbf{Process}              & \textbf{Computation time (in seconds)} \\
\hline
Mask Generation & 0.015                                                \\
\hline
Resampling and DON   & 0.153                                                \\
\hline
Clustering           & 0.19                                                 \\
\hline
Plane Segmentation   & 0.0831                                               \\
\hline
Filtering            & 0.301                                                \\
\hline
6DOF Pose Estimation & 0.159                                                \\
\hline
\textbf{Total Time}           & \textbf{0.901}                                       \\
\hline
\end{tabular} 
\end{adjustbox}
\caption{Results are averaged over 25 samples.}
\end{table}

Table 3 shows the total processing time for both box segmentation and box localization modules. The total processing time comes out to be 0.9 seconds. These results are comparable to the results demonstrated in \cite{Mahler and Goldberg}. As opposed to \cite{Mahler and Goldberg}, which requires very high computation resources to generate pick points, our approach works very well on a regular laptop. This makes our approach cost-effective. In addition, our approach is easily scalable to multiple boxes since it does not require any prior knowledge of the boxes or any form of training.

\section{Conclusion and Future Work}
We presented a dual sensor system for box segmentation and box location computation in bin picking applications. Two algorithms for box mask segmentation and box localization are proposed. The box mask segmentation algorithm is found to be robust in case of no texture, minimal and high texture boxes. However, the performance for the glossy surface is limited by 3D depth sensor measurement. The unified vision algorithm is capable of computing accurate poses in all three axes with an average error of +/- 3.3mm; however, the picking operations were unaffected due to the error in z-axis because of the tool with single suction type vacuum gripper which has a tolerance of up to +/- 5 mm. The real-time experimental demonstrates a consistent detection time of approximately 1 second on the specified hardware and test cases. We have empirically proved the feasibility of the proposed algorithms for bin-picking application and it can be adopted for many real-world box picking operations in warehouses and e-commerce industry. This solution has the potential to be extended for other objects such as cylinders using shape decomposition algorithms.

\end{document}